\documentclass[11pt]{article}

\usepackage[preprint]{acl}

\usepackage{times}
\usepackage{latexsym}
\usepackage{booktabs}
\usepackage[T1]{fontenc}

\usepackage[utf8]{inputenc}

\usepackage{microtype}

\usepackage{inconsolata}

\usepackage{graphicx}

\usepackage{amssymb}
\usepackage{amsmath}
\usepackage{amsthm}
\usepackage{enumitem}
\usepackage{subcaption}
\def\Algnameabbr{\texttt{ADS}}
\def\Algname{\textbf{A}daptive \textbf{D}ata \textbf{S}cheduling}

\usepackage{algorithm}
\usepackage{graphicx}
\usepackage{caption}
\usepackage{wrapfig}
\usepackage{algorithmic}
\usepackage{ulem}
\DeclareMathOperator*{\argmin}{arg\,min}
\DeclareMathOperator*{\argmax}{arg\,max}
\newtheorem{proposition}{Proposition}[section]

\title{Learning at the Right Pace: Adaptive Data Scheduling \\ Improves LLM Reinforcement Learning}

\author{
 \textbf{Zicheng Xu\textsuperscript{1}\thanks{Equal contribution, ordered by rolling dices.}},
 \textbf{Ruixuan Zhang\textsuperscript{1}\footnotemark[1]},
 \textbf{Yu-Neng Chuang\textsuperscript{2}},
 \textbf{Xiuyi Lou\textsuperscript{1}},
 \textbf{Hoang Anh Duy Le\textsuperscript{2}},
 \\
 \textbf{Oren Gal\textsuperscript{3}},
 \textbf{Alexander Szalay\textsuperscript{1}},
 \textbf{Zhaozhuo Xu\textsuperscript{4}},
 \textbf{Guanchu Wang\textsuperscript{5}${}^\dagger$},
 \textbf{Vladimir Braverman\textsuperscript{1}\thanks{Correspondence to: Vladimir Braverman \href{mailto:vova@cs.jhu.edu}{vova@cs.jhu} \\ \href{mailto:vova@cs.jhu.edu}{.edu} and Guanchu Wang \href{mailto:gwang16@charlotte.edu}{gwang16@charlotte.edu}}} 
\\
\\
 \textsuperscript{1}Johns Hopkins University,
 \textsuperscript{2}Rice University,
 \textsuperscript{3}University of Haifa,
 \\
 \textsuperscript{4}Workato,
 \textsuperscript{5}University of North Carolina at Charlotte
}

\begin{document}
\maketitle
\vspace{-5mm}
\begin{abstract}
Large Language Models (LLMs) achieve remarkable reasoning capabilities through reinforcement learning (RL) post-training. 
However, existing RL post-training commonly relies on uniform data sampling, which ignores the semantic structure of the training data and the evolving capability of the training policy. 
To address these limitations, we propose \Algname{} (\Algnameabbr{}), a dual-level data scheduling framework that combines adaptive sampling across semantic clusters with policy-boundary sample selection.
At the cluster level, \Algnameabbr{} organizes samples according to semantic patterns and adapts the inter-cluster sampling distribution to prioritize clusters within the policy's current reach.
At the sample level, \Algnameabbr{} tracks prompts near the policy's capability boundary within each cluster to obtain informative relative advantages. 
Experimental results across three LLMs and seven reasoning benchmarks demonstrate that \Algnameabbr{} improves average accuracy by \textbf{5.5\%} over Group Relative Policy Optimization (GRPO).
Notably, \Algnameabbr{} consistently improves RL methods with different objective designs, highlighting its potential as a general data scheduling strategy. The source code is available at: \url{https://github.com/Richard-zrx/ADS}.
\end{abstract}

\section{Introduction}

Recent advancements in reinforcement learning (RL) post-training for large language models (LLMs) have demonstrated impressive improvements in reasoning capabilities, exemplified by reasoning models such as DeepSeek-R1~\cite{guo2025deepseek} and Qwen3~\cite{yang2025qwen3}. Unlike supervised fine-tuning, RL post-training enables LLMs to improve through reward-driven exploration, allowing them to develop more effective reasoning behaviors through trial-and-error optimization. Among these methods, Group Relative Policy Optimization (GRPO)~\cite{guo2025deepseek} has become a widely adopted RL paradigm, in which multiple rollouts are sampled for each training sample, and the policy is optimized based on their relative outcome rewards. Complementary advancements such as On-Policy Distillation (OPD)~\cite{agarwal2024policy}, Dynamic sAmpling Policy Optimization (DAPO)~\cite{yu2025dapo}, and Group Sequence Policy Optimization (GSPO)~\cite{zheng2025group} further improve LLM post-training.

\begin{figure}
    \centering
    \includegraphics[width=\linewidth]{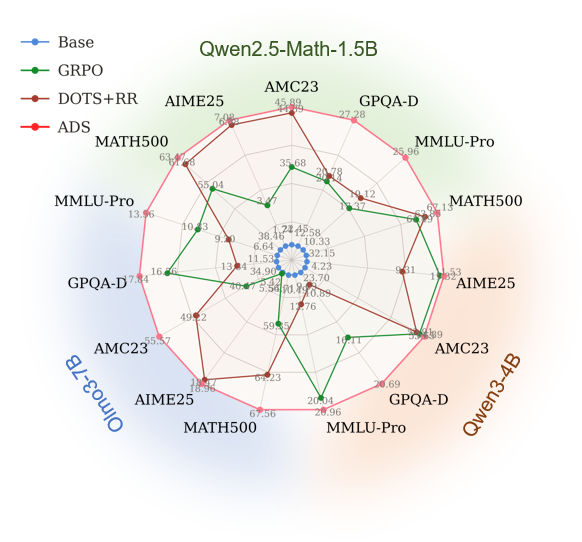}
    \caption{Comprehensive performance of \Algnameabbr{} on three LLMs and five reasoning datasets.}
    \label{fig:radar}
    \vspace{-3mm}
\end{figure}

However, current RL-based post-training commonly relies on an unstructured and ineffective data sampling scheme~\cite{liu2025understanding, sun2025improving, tian2021independent}. In practice, samples are typically drawn from the full training set according to a uniform distribution, treating the dataset as an undifferentiated pool. This uniform data schedule exhibits two fundamental limitations that hinder robust policy optimization. 
First, uniform sampling ignores the semantic structure of the training data.
Semantically related samples often share reasoning patterns, while the policy's ability to learn different patterns can develop at varying paces.
However, uniform sampling allocates training effort across these semantic patterns according to their prevalence in the dataset, without considering which are currently within the policy's reach.
This misses opportunities to consolidate progress on learnable patterns through related samples and adjust the training focus as the policy improves.
Second, existing works~\cite{klink2022curriculum,zhan2025exgrpo} have shown that not all samples are equally useful for RL policy optimization. 
In GRPO-style training, samples near the policy's current capability boundary are especially informative since they produce rollout groups with both correct and incorrect responses, yielding meaningful relative advantages.
Uniform sampling does not account for where samples lie relative to this boundary and may select samples that are either too easy or too difficult, causing advantage collapse and ineffective policy updates.
Together, these limitations motivate a structured data schedule that complements prompt-level selection~\cite{yu2025dapo,sun2025improving} with adaptive allocation across semantic patterns.

To address these limitations, we propose \textbf{A}daptive \textbf{D}ata \textbf{S}cheduling (\Algnameabbr{}), a dual-level scheduling framework that paces RL post-training according to both semantic structure and policy capability. 
Rather than treating all prompts as independent samples, \Algnameabbr{} first organizes the training data into semantic clusters according to shared patterns and meanings. 
During training, it maintains an adaptive inter-cluster distribution to control how frequently each semantic cluster enters the current training schedule, selecting ones that are within reach under the current policy capability.  
This cluster-level schedule consolidates progress on semantically related samples while adapting the training focus to the policy's evolving capabilities.
Within each selected semantic cluster, \Algnameabbr{} further performs sample-level intra-cluster scheduling to continuously keep the sampled prompts close to the policy's current capability boundary.
As the policy improves, this sample-level schedule shifts with the policy boundary, preventing training from being dominated by samples that have become too easy or remain too difficult.

Empirically, we evaluate \Algnameabbr{} on seven mathematical and out-of-distribution (OOD) scientific reasoning benchmarks and three LLMs. Across all settings, \Algnameabbr{} improves average accuracy by \textbf{5.5\%} over GRPO,  with Figure~\ref{fig:radar} visualizing the gains on five representative benchmarks. Our contributions are summarized as follows:
\begin{itemize}[leftmargin=10pt]
    \itemsep=-2pt
    \item \textbf{Dual-Level Adaptive Scheduling:} \Algnameabbr{} adaptively prioritizes semantic clusters within the policy's reach and tracks policy-boundary samples within each cluster to obtain informative updates.
    \item \textbf{RL Objective-Agnostic:} \Algnameabbr{} optimizes only the data sampling distribution without altering the underlying RL objective, allowing it to improve methods with different RL objective designs.
    \item \textbf{Evaluation:} Across three LLMs and seven reasoning benchmarks, \Algnameabbr{} consistently improves reasoning accuracy against baselines.
\end{itemize}

\section{Preliminary}

\subsection{Notations}

We consider a parameterized LLM policy $\pi_\theta$ in this work, where $\theta$ represents the trainable parameters. 
Let $\mathcal{D}=\{(x_i, y_i)\}_{i=1}^N$ denote the training dataset, where each $x_i$ represents a sample prompt and $y_i$ is its corresponding reference solution.  
In this work, we aim to construct a data schedule over $\mathcal{D}$ that determines which samples should be used at each stage of RL post-training. The schedule is designed to account for both the semantic structure of $\mathcal{D}$ and current capability of $\pi_\theta$ such that policy optimization receives more informative training samples.

\subsection{Group Relative Policy Optimization}
\label{sec:grpo_prelim}
Group Relative Policy Optimization (GRPO) has emerged as a widely used RL with verifiable reward framework for post-training LLMs~\cite{guo2025deepseek, zhang2025survey}. Given a prompt $x_i$, GRPO samples a group of $G$ responses $\{o_{i,j}\}_{j=1}^{G}$ from the policy $\pi_\theta$. Each response rollout is assigned a sequence-level reward $\mathcal{R}(x_i, o_{i,j})$, which is typically binary for correctness. To optimize the policy without relying on a separate memory-intensive value network, GRPO utilizes relative outcome comparisons within each group. Specifically, it computes a normalized advantage for each trajectory: $A_{i,j} = (\mathcal{R}(x_i, o_{i,j}) - \mu_i) / (\sigma_i + \epsilon)$, where $\mu_i$ and $\sigma_i$ are the empirical mean and standard deviation of the $G$ rewards. The policy is then updated to increase the likelihood of trajectories with positive advantages and vice versa.

While GRPO provides a highly efficient objective for policy optimization, the framework relies on a naive, unstructured data sampling strategy. Formally, the data schedule is static and can be represented as maximizing the expected loss over sample prompts drawn uniformly from the dataset at any given step:

\begin{equation}
    \max_{\theta} \mathbb{E}_{x_i \sim \mathcal{U}(\mathcal{D})} \left[ \mathcal{L}_{\text{GRPO}}(\pi_\theta; x_i) \right],
\end{equation}
where $\mathcal{U}(\mathcal{D})$ denotes the uniform distribution over the entire training set $\mathcal{D}$, and $\mathcal{L}_{\text{GRPO}}$ represents the GRPO objective\footnote{The full GRPO loss function can be referred to Equation (1) in the DeepSeek-R1 technical report~\cite{guo2025deepseek}.}. 

This uniform schedule assumes all samples are equally informative for optimization at any given step, regardless of their semantic structure or their difficulty under the current policy. 
However, prior works~\cite{klink2022curriculum,zhan2025exgrpo} have demonstrated that this assumption is flawed and RL benefits most from training on samples near the policy's capability boundary. 
In GRPO, overly easy samples often yield all-correct rollout groups, while overly difficult prompts often yield all-incorrect groups. In both cases, the normalized advantages collapse toward zero, providing no useful gradient for policy improvement. Informative updates arise when a prompt produces both successful and failed rollouts within the same group, which occurs near the policy's current capability boundary. We provide a theoretical justification for this boundary criterion in Appendix~\ref{sec:5050just}. This limitation motivates our adaptive scheduling framework, which preserves the GRPO objective but replaces the static uniform sampling with a dual-level data schedule that adapts to both the data semantic structure and policy capability boundary.

\section{\Algname{}}
\begin{figure*}
    \centering
    \captionsetup{belowskip=-3mm}
    \includegraphics[width=1.0\linewidth]{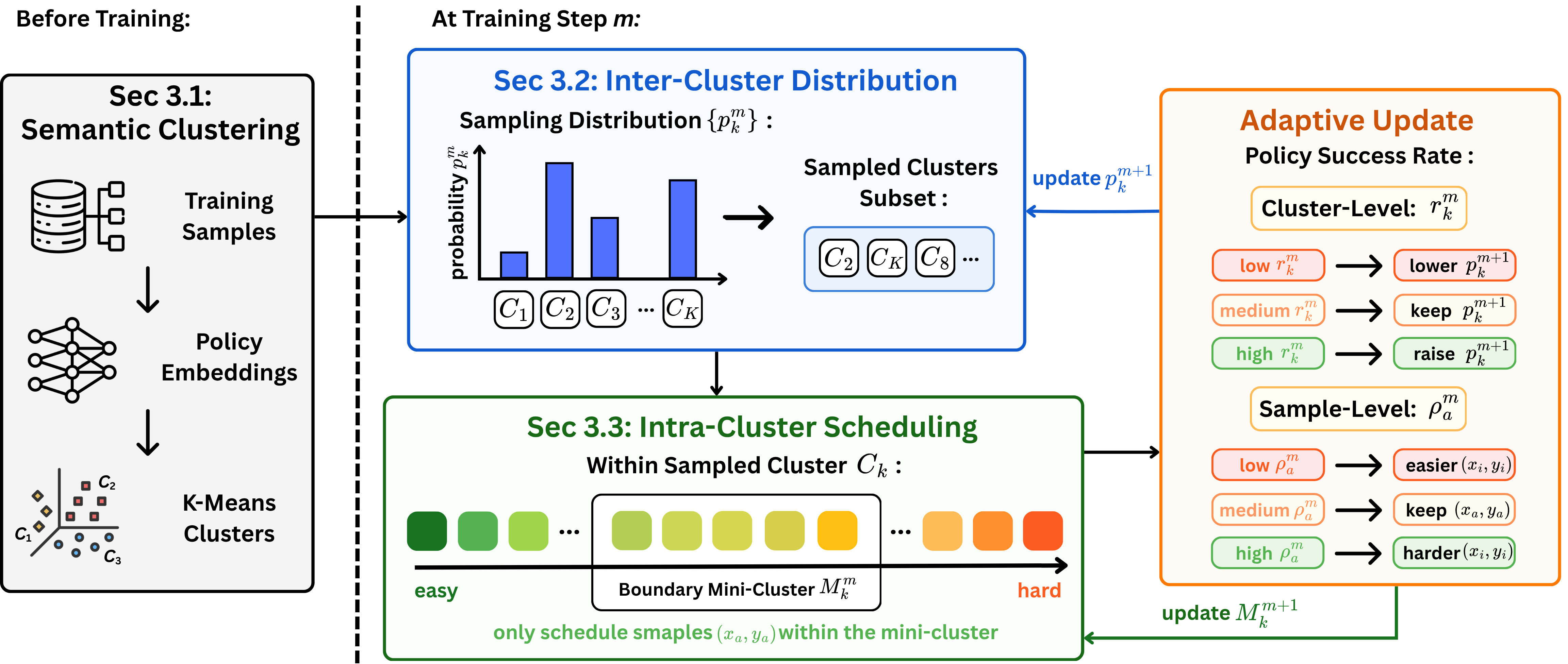}
    \caption{The overall framework of \Algnameabbr{}.}
    \label{fig:framework}
\end{figure*}

In this section, we introduce \Algnameabbr{}, a dual-level adaptive data scheduling framework that selects samples according to the capability of the policy. The overall framework of \Algnameabbr{} is illustrated in Figure~\ref{fig:framework}. \Algnameabbr{} consists of three key modules: semantic clustering, inter-cluster sampling distribution, and intra-cluster scheduling. Together, these modules organize the dataset into coherent semantic clusters and generate an optimal schedule for training samples. 

\subsection{Semantic Clustering}
To provide a structured view of the training dataset, \Algnameabbr{} first organizes training samples into semantic clusters. Each cluster has unified semantic patterns, allowing \Algnameabbr{} to control the data schedule at the cluster level rather than treating the dataset as an undifferentiated pool of independent samples. To construct these semantic clusters, \Algnameabbr{} leverages the base policy's representation space, which captures both the semantic content and structural patterns of each sample. Clustering in this space produces semantically coherent groups, providing the foundation for the later dual-level data schedule.

Formally, for each training sample $(x_i, y_i) \in \mathcal{D}$, \Algnameabbr{} encodes the concatenated sequence $\boldsymbol{z}_i = x_i \oplus y_i$ with the base policy $\pi_\theta$. To capture the semantic patterns of the sequence, \Algnameabbr{} extracts the final-layer hidden states $\tilde{\pi}_\theta(\cdot)$ at each token position and applies mean pooling over these representations:
\begin{equation}
    \mathbf{e}_i = \frac{1}{|\boldsymbol{z}_i|} \sum_{t = 1}^{|\boldsymbol{z}_i|} \tilde{\pi}_\theta(\boldsymbol{z}_{i,[1:t]}).
    \label{eq:embedding}
\end{equation}
Given the resulting set of embeddings $\{\mathbf{e}_i\}_{i=1}^N$, \Algnameabbr{} applies K-Means clustering to partition the dataset into $K$ distinct semantic clusters:
\begin{equation}
    \mathcal{C} =\{C_1, C_2, \dots, C_K\} = \mathrm{KMeans}(\{\mathbf{e}_i\}_{i=1}^N; K).
    \label{eq:semantic_clustering}
\end{equation}
By partitioning the representation space, each resulting cluster $C_k$ forms a cohesive semantic domain instead of discrete independent samples. The resulting cluster set $\mathcal{C}$ supports the subsequent dual-level schedule, with inter-cluster sampling adapting the training distribution across semantic clusters and intra-cluster scheduling selecting capability-matched samples within each cluster.
\subsection{Inter-Cluster Distribution}
\label{sec:inter-cluster}

During training, \Algnameabbr{} actively selects clusters that align with the current capability of the training policy to consolidate progress on related samples.
Specifically, at training step $m$, let $p_k^m$ denote the probability of adding cluster $C_k$ to the training data, initialized uniformly with $p_k^0 = 1/K$.
As training proceeds, \Algnameabbr{} adapts this inter-cluster distribution according to the current policy's capability on each cluster. 
To estimate this capability, \Algnameabbr{} maintains a cluster-level success rate $r_k^m \in [0,1]$ for each cluster $C_k$. When $C_k$ is sampled at step $m$, $r_k^m$ is computed as the fraction of correct rollouts among all rollouts generated from samples in $C_k$; otherwise, the previous success rate is carried forward.
Clusters with higher success rates are more within reach for the current policy and should be prioritized to solidify current progress with larger sampling probabilities. Clusters with low success rates are beyond the policy's current capability and are temporarily down-weighted to avoid sampling clusters that produce sparse advantages. 
Formally, \Algnameabbr{} updates the probability of selecting each cluster as:
\begin{equation}
    p_k^{m+1}
    =
    \frac{r_k^m}{\sum_{j=1}^{K} r_j^m}.
    \label{eq:inter_cluster_probability}
\end{equation}
In practice, \Algnameabbr{} gradually adapts the cluster-level distribution with exponential smoothing, preventing sudden shifts caused by noisy rollout outcomes. The updated distribution is then used to sample clusters that enter the subsequent policy update. 
In this way, the inter-cluster distribution adapts the training schedule across semantic patterns according to the policy's observed success. Complementing this allocation, intra-cluster scheduling tracks policy-boundary prompts within each selected cluster to obtain informative relative advantages.

\subsection{Intra-Cluster Scheduling} 

Given the selected clusters by the inter-cluster distribution, \Algnameabbr{} further performs intra-cluster scheduling during training to select optimal policy-boundary samples.

\paragraph{Policy-Boundary Samples.}
As discussed in Section~\ref{sec:grpo_prelim}, GRPO-style training relies on contrast within each rollout group.  
The most informative samples are those near the policy's current capability boundary, where the task is neither already mastered nor entirely out of reach~\cite{klink2022curriculum,zhan2025exgrpo}. 
We refer to these samples as policy-boundary samples.
Under standard binary outcome rewards, policy-boundary samples correspond to empirical success rates near $50\%$, where rollout groups are likely to contain both correct and incorrect responses.
However, cluster-level success alone cannot identify these samples, since samples within the same semantic cluster can differ substantially in difficulty.
Their success rates also shift with each policy update, such that samples near the boundary at one stage may move away from it later.
Thus, \Algnameabbr{} employs intra-cluster scheduling to continuously evaluate samples within each cluster and identify those that remain near the policy's current capability boundary.

\paragraph{Selecting Policy-Boundary Samples.} 

To track policy-boundary samples efficiently, \Algnameabbr{} maintains a boundary mini-cluster $M_k^m \subseteq C_k$ for each semantic cluster $C_k$ at each step.
The mini-cluster serves as the policy-boundary candidate set within $C_k$: when $C_k$ is selected by the inter-cluster sampling distribution, \Algnameabbr{} uses only the samples in $M_k^m$ for the current policy update and adapts this set according to rollout feedback.
To make this adaptation efficient and differentiate samples within each cluster, \Algnameabbr{} first assigns each sample a fixed offline difficulty score.
Specifically, before training, \Algnameabbr{} assigns each sample $(x_i,y_i)\in C_k$ a difficulty score using the negative log-likelihood of the reference solution under the base policy:
\begin{equation}
    d_i = -\frac{1}{|y_i|}\sum_{t=1}^{|y_i|}
    \log \pi_\theta(y_{i,t}\mid x_i,y_{i,<t}).
    \label{eq:difficulty}
\end{equation}
A lower score indicates that the solution $y_i$ is more likely under the base policy given prompt $x_i$, so the corresponding sample is treated as easier, while a higher score sample is treated as harder.
Within each cluster, \Algnameabbr{} orders samples by this difficulty score and initializes $M_k^0$ with the easiest samples in $C_k$ as a conservative warm start, since no online rollout feedback is available before training.

As training proceeds, this initial boundary mini-cluster is adapted using rollout feedback so that it increasingly contains policy-boundary samples. For each candidate sample $(x_a,y_a)\in M_k^m$, let $\rho_a^m \in [0,1]$ denote the empirical success rate of the rollout group generated from $x_a$ at step $m$.
Since standard outcome rewards are binary, policy-boundary sample corresponds to a success rate of near $50\%$, while lower and higher success rates indicate that the sample is too difficult or too easy for the current policy, respectively.
Enforcing an exact $50\%$ success rate would be overly restrictive and \Algnameabbr{} uses a tolerance $\epsilon$ and keeps samples whose success rates fall within $[0.5-\epsilon,0.5+\epsilon]$ to identify policy-boundary samples.
Samples outside this band are replaced by nearby alternatives from the same semantic cluster according to the offline difficulty ordering.
Formally, for each candidate sample $(x_a,y_a)\in M_k^m$ with offline difficulty score $d_a$, we define the sample-level update rule as
\begin{equation}
    \phi^m(x_a,y_a) =
    \begin{cases}
        (x_a,y_a),
        & |\rho_a^m-0.5| \leq \epsilon, \\[4pt]
        \displaystyle
        \argmax_{\substack{(x_i,y_i)\in C_k \\ d_i < d_a}} d_i,
        & \rho_a^m < 0.5-\epsilon, \\[4pt]
        \displaystyle
        \argmin_{\substack{(x_i,y_i)\in C_k \\ d_i > d_a}} d_i,
        & \rho_a^m > 0.5+\epsilon.
    \end{cases}
    \label{eq:sample_update_rule}
\end{equation}
The boundary mini-cluster is then updated by applying this rule to all candidate samples:
\begin{equation}
    M_k^{m+1} = \{\phi^m(x_a,y_a) ~|~ (x_a,y_a)\in M_k^m\}.
    \label{eq:mini_cluster_update}
\end{equation}
This local replacement rule allows the boundary mini-cluster to track policy-boundary samples efficiently.
To encourage coverage across clusters, \Algnameabbr{} excludes a cluster from further inter-cluster sampling once all its samples have been scheduled at least once by the intra-cluster schedule.
Sampling probabilities are then renormalized over the remaining eligible clusters.
At the start of the next epoch, all clusters are reactivated.
Overall, intra-cluster scheduling combines offline difficulty ordering with online rollout feedback to adaptively schedule policy-boundary samples to enhance training effectiveness.
We provide the full algorithm of \Algnameabbr{} in Appendix~\ref{sec:algo}.

\section{Experiments}
\label{sec:exp}
In this section, we conduct experiments to evaluate the performance of \Algnameabbr{} framework, aiming to answer the following research questions: \textbf{RQ1:} Does \Algnameabbr{} improve LLM post-training to produce more accurate reasoning? \textbf{RQ2:} Can \Algnameabbr{} generalize its gains across different RL post-training objectives beyond GRPO? \textbf{RQ3:} How sensitive is \Algnameabbr{} to the policy used for semantic clustering?

\subsection{Experimental Setup}
\label{sec:setup}
We specify the training data, models, evaluation datasets, and baseline methods below. We provide more information onimplementation and evaluation details in Appendix~\ref{sec:implementation_details} and Appendix~\ref{app:eval_detail}.

\noindent\textbf{Training.} We build our training data from the OpenR1-Math-220k\footnote{\url{https://huggingface.co/datasets/open-r1/OpenR1-Math-220k}} dataset, following the filtering procedure of~\citet{yan2025learning}, resulting in 11.5k high quality prompts with solutions.

\noindent\textbf{Models.} We evaluate \Algnameabbr{} with three representative LLMs: Qwen2.5-1.5B-Math~\cite{yang2024qwen2}, Qwen3-4B-Base~\cite{yang2025qwen3}, and Olmo3-7B-SFT~\cite{olmo2025olmo}.

\noindent\textbf{Evaluation.}
We evaluate \Algnameabbr{} on seven widely used reasoning benchmarks. For mathematical reasoning, we evaluate on five standard mathematics benchmarks: AIME24, AIME25, AMC23~\cite{aime24, aime25}, MATH-500~\cite{lightman2023lets}, and Minerva~\cite{lewkowycz2022solving}. To assess out-of-distribution (OOD) generalization beyond mathematics, we further evaluate on two scientific reasoning benchmarks, GPQA-Diamond~\cite{rein2023gpqa} and MMLU-Pro~\cite{wang2024mmlu}. We report Mean@16 accuracy(\%) as the primary evaluation metric, with standard deviation from three independent runs. 
We train until convergence and select the checkpoint with the highest validation accuracy on GSM8K~\cite{gsm8k}.

\noindent\textbf{Baseline Methods.}
\textbf{Base Model:} We evaluate each LLM under its default inference setting without any post-training.
\textbf{GRPO:} GRPO~\cite{guo2025deepseek} samples training prompts uniformly from the full dataset and optimizes the policy using group-relative advantages computed from multiple rollouts to the same prompt.
\textbf{DOTS+RR:} DOTS+RR~\cite{sun2025improving} extends GRPO with difficulty-targeted online data selection and rollout replay. It estimates which prompts are likely useful and reuses rollouts from a replay buffer.

\providecommand{\std}[1]{{\scriptsize$\pm$#1}}
\begin{table*}[t]
\centering
\small
\setlength{\tabcolsep}{2.5pt}
\renewcommand{\arraystretch}{1.08}
\resizebox{\textwidth}{!}{
\begin{tabular}{l|l|cccccccc}
\toprule
Method & Metric & MATH-500 & AIME 25 & AIME 24 & Minerva & AMC 23 & GPQA-D & MMLU-Pro & Average \\
\midrule

\multicolumn{10}{c}{\textbf{Qwen2.5-Math-1.5B}} \\
\midrule
Base 
& Mean@16 & 38.46\std{0.29} & 1.74\std{0.10} & 2.64\std{0.40} & 8.90\std{0.48} & 22.45\std{1.69} & 12.58\std{0.26} & 10.33\std{0.19} & 13.87 \\
& Pass@16 & 80.60\std{0.57} & 13.33\std{0.00} & 24.45\std{5.66} & 28.80\std{0.62} & 80.83\std{2.36} & 80.13\std{1.95} & 59.13\std{1.09} & 52.47 \\

GRPO 
& Mean@16 & 55.04\std{0.08} & 3.47\std{0.43} & 5.63\std{1.03} & 13.63\std{0.13} & 35.68\std{0.60} & 20.14\std{0.53} & 17.37\std{0.13} & 21.57 \\
& Pass@16 & 82.87\std{0.90} & 20.00\std{4.71} & \textbf{34.44\std{4.16}} & 31.86\std{0.92} & 79.17\std{4.25} & \underline{87.03\std{1.26}} & \underline{69.80\std{0.57}} & 57.88 \\

DOTS+RR 
& Mean@16 & \underline{61.68\std{0.15}} & \underline{6.88\std{0.00}} & \underline{8.75\std{0.34}} & \underline{14.79\std{0.05}} & \underline{44.89\std{1.18}} & \underline{20.78\std{0.88}} & \underline{19.12\std{0.31}} & \underline{25.27} \\
& Pass@16 & \underline{83.33\std{0.90}} & \textbf{33.33\std{2.72}} & \underline{33.33\std{0.00}} & \underline{32.36\std{0.52}} & \underline{83.33\std{3.12}} & 77.61\std{1.86} & 68.33\std{0.57} & \underline{58.80} \\

\Algnameabbr{}
& Mean@16 & \textbf{63.47\std{0.30}} & \textbf{7.08\std{0.34}} & \textbf{9.44\std{1.13}} & \textbf{17.20\std{0.22}} & \textbf{45.89\std{1.74}} & \textbf{27.28\std{0.84}} & \textbf{25.96\std{0.08}} & \textbf{28.05} \\
& Pass@16 & \textbf{84.80\std{0.43}} & \underline{30.00\std{4.71}} & 32.22\std{5.66} & \textbf{33.21\std{0.17}} & \textbf{87.50\std{5.40}} & \textbf{89.39\std{0.41}} & \textbf{74.73\std{0.68}} & \textbf{61.69} \\

\midrule

\multicolumn{10}{c}{\textbf{Qwen3-4B-Base}} \\
\midrule
Base 
& Mean@16 & 32.15\std{0.79} & 4.23\std{0.49} & 5.83\std{0.90} & 10.27\std{0.20} & 23.70\std{0.71} & 9.69\std{0.23} & 10.49\std{0.27} & 13.77 \\
& Pass@16 & 79.87\std{0.84} & 27.78\std{3.14} & 27.78\std{3.14} & 35.05\std{0.62} & 78.33\std{2.36} & 62.79\std{1.72} & 59.93\std{1.09} & 53.08 \\

GRPO 
& Mean@16 & 61.49\std{0.05} & \underline{11.32\std{0.64}} & 12.22\std{0.78} & 18.40\std{0.10} & \underline{53.65\std{0.27}} & \underline{16.11\std{0.65}} & \underline{20.04\std{0.49}} & \underline{27.60} \\
& Pass@16 & \textbf{\underline{84.73\std{0.94}}} & \textbf{\underline{37.78\std{1.57}}} & 31.11\std{3.14} & 39.58\std{0.62} & \textbf{\underline{89.17\std{3.12}}} & \underline{69.70\std{2.18}} & \textbf{65.47\std{1.18}} & \underline{59.65} \\

DOTS+RR 
& Mean@16 & \underline{63.85\std{0.18}} & 9.31\std{0.43} & \textbf{14.09\std{0.55}} & \textbf{21.35\std{0.12}} & 52.91\std{0.29} & 10.89\std{0.12} & 12.76\std{0.14} & 26.45 \\
& Pass@16 & 84.00\std{0.43} & 30.00\std{4.71} & \textbf{34.45\std{3.14}} & \underline{40.19\std{0.17}} & \textbf{\underline{89.17\std{2.36}}} & 59.09\std{0.41} & 57.27\std{0.34} & 56.31 \\

\Algnameabbr{}
& Mean@16 & \textbf{67.13\std{0.08}} & \textbf{11.53\std{0.49}} & \underline{12.50\std{1.03}} & \underline{20.89\std{0.10}} & \textbf{54.89\std{0.39}} & \textbf{20.69\std{0.45}} & \textbf{20.96\std{0.20}} & \textbf{29.80} \\
& Pass@16 & \textbf{\underline{84.73\std{0.19}}} & \textbf{\underline{37.78\std{1.57}}} & \underline{33.33\std{0.00}} & \textbf{40.69\std{0.91}} & \textbf{\underline{89.17\std{1.18}}} & \textbf{71.55\std{0.86}} & \underline{63.80\std{0.75}} & \textbf{60.15} \\

\midrule

\multicolumn{10}{c}{\textbf{Olmo3-7B-SFT}} \\
\midrule
Base 
& Mean@16 & 54.71\std{0.21} & 5.42\std{0.29} & 4.31\std{0.10} & 16.37\std{0.16} & 34.90\std{0.97} & 11.53\std{0.64} & 6.64\std{0.09} & 19.13 \\
& Pass@16 & 82.67\std{0.34} & 27.78\std{3.14} & 24.44\std{4.16} & 37.13\std{1.08} & 80.83\std{1.18} & 56.91\std{1.72} & 43.00\std{1.23} & 50.39 \\

GRPO 
& Mean@16 & 59.35\std{0.19} & 5.56\std{0.49} & 6.67\std{0.45} & \underline{17.21\std{0.26}} & 40.47\std{2.27} & \underline{16.56\std{0.44}} & \underline{10.83\std{0.15}} & 22.38 \\
& Pass@16 & 84.27\std{0.74} & 24.44\std{1.57} & 33.33\std{2.72} & \underline{37.38\std{1.21}} & 84.17\std{4.71} & \textbf{69.53\std{1.32}} & \underline{52.60\std{0.65}} & 55.10 \\

DOTS+RR 
& Mean@16 & \underline{64.23\std{0.04}} & \underline{18.47\std{0.39}} & \underline{15.28\std{1.61}} & 17.16\std{0.08} & \underline{49.22\std{0.97}} & 13.34\std{0.09} & 9.20\std{0.08} & \underline{26.70} \\
& Pass@16 & \underline{86.27\std{0.09}} & \underline{44.45\std{3.14}} & \underline{50.00\std{4.71}} & 34.68\std{1.48} & \underline{85.83\std{3.12}} & 63.47\std{2.27} & 51.87\std{0.66} & \underline{59.51} \\

\Algnameabbr{}
& Mean@16 & \textbf{67.56\std{0.16}} & \textbf{18.96\std{0.30}} & \textbf{18.33\std{1.94}} & \textbf{18.58\std{0.17}} & \textbf{55.57\std{1.25}} & \textbf{17.84\std{0.11}} & \textbf{13.56\std{0.15}} & \textbf{30.06} \\
& Pass@16 & \textbf{88.20\std{0.71}} & \textbf{45.55\std{3.14}} & \textbf{54.45\std{3.14}} & \textbf{37.50\std{0.79}} & \textbf{95.00\std{0.00}} & \underline{69.36\std{1.04}} & \textbf{59.53\std{0.41}} & \textbf{64.23} \\

\bottomrule
\end{tabular}
}
\caption{Results on seven mathematical and OOD scientific reasoning benchmarks and three representative LLMs.}
\label{tab:main_results}
\vspace{-3mm}
\end{table*}

\noindent\textbf{Implementation Details.}
We employ the verl framework~\cite{sheng2025hybridflow} to perform all training. For all baseline training methods and \Algnameabbr{}, we set the rollout batch size to 128, group size to 8, sampling temperature to 1.0, learning rate to $1\times 10^{-6}$, and maximum generation length to $8192$. For \Algnameabbr{}, we set the number of clusters $K=64$, mini-cluster size $B=32$, and a policy-boundary band $[0.33, 0.67]$ with $\epsilon=0.17$.

\subsection{Performance on Reasoning Tasks~(RQ1)}
Table~\ref{tab:main_results} shows the Mean@16 and Pass@16 accuracy(\%) of \Algnameabbr{} compared with baseline methods.

\vspace{1mm}
\noindent\textbf{Accuracy Improvement.} As shown in Table~\ref{tab:main_results}, \Algnameabbr{} consistently outperforms baseline methods across different backbone models and reasoning benchmarks. The improvements are reflected in both Mean@16 and Pass@16, which demonstrates that \Algnameabbr{} enhances both average correctness and multi-sample exploration ability of the LLMs. These results show that replacing uniform data sampling with an adaptive schedule that accounts for both the semantic structure and the policy capability boundary potentially provides more effective training signals for RL-based post-training.

\vspace{1mm}
\noindent\textbf{OOD Generalization.} \Algnameabbr{} also demonstrates strong OOD generalization beyond the mathematical reasoning domain used for training. Specifically, \Algnameabbr{} exhibits strong and stable accuracy gains in both GPQA-D and MMLU-Pro datasets compared to baseline methods in Table~\ref{tab:main_results}. The domains from these datasets span biology, physics, chemistry, computer science, business, etc. These robust gains suggest that \Algnameabbr{} does not simply overfit to patterns in mathematical problems, but rather strengthens general reasoning behaviors through generating an adaptive data schedule.

\vspace{1mm}
\noindent\textbf{Model Scale Consistency.} \Algnameabbr{} achieves consistent improvements across models of different parameter scales. Table~\ref{tab:main_results} shows evaluation results on 1.5B, 4B, and 7B LLMs, and \Algnameabbr{} improves reasoning accuracy across all scales. Table~\ref{tab:qwen3_scaling} in Appendix~\ref{sec:scaling} further supports this finding through a controlled comparison of Qwen3-Base models at 1.7B, 4B, and 8B. This consistency is supported by the adaptive design of \Algnameabbr{}, which updates the data schedule from each policy's own rollout feedback and adjusts to different model capabilities.

\begin{table}[t]
\centering
\small
\setlength{\tabcolsep}{4pt}
\renewcommand{\arraystretch}{1.08}
\begin{tabular}{@{}lrrrrrr@{}}
\toprule
& \multicolumn{2}{c}{\textbf{GSPO}}
& \multicolumn{2}{c}{\textbf{DAPO}}
& \multicolumn{2}{c}{\textbf{OPD}} \\
\cmidrule(lr){2-3}
\cmidrule(lr){4-5}
\cmidrule(lr){6-7}
Dataset
& Orig. & +\Algnameabbr{}
& Orig. & +\Algnameabbr{}
& Orig. & +\Algnameabbr{} \\
\midrule
MATH-500
& 62.53 & \textbf{65.61}
& 65.05 & \textbf{66.62}
& 79.22 & \textbf{79.81} \\
AIME 25
& 7.08 & \textbf{9.37}
& 7.85 & \textbf{9.31}
& 35.07 & \textbf{36.25} \\
AIME 24
& 9.59 & \textbf{10.00}
& \textbf{11.53} & 10.21
& 44.51 & \textbf{46.46} \\
Minerva
& 14.92 & \textbf{15.85}
& 16.68 & \textbf{17.75}
& 22.78 & \textbf{23.94} \\
AMC 23
& 45.41 & \textbf{48.91}
& 47.87 & \textbf{49.01}
& 85.68 & \textbf{87.50} \\
GPQA-D
& 21.83 & \textbf{24.43}
& 23.71 & \textbf{26.73}
& 37.11 & \textbf{38.67} \\
MMLU-Pro
& 20.42 & \textbf{24.88}
& 23.82 & \textbf{26.59}
& 36.28 & \textbf{38.46} \\
\bottomrule
\end{tabular}
\caption{Mean@16 accuracy (\%) of \Algnameabbr{} across three different RL post-training objectives. GSPO and DAPO are performed on Qwen2.5-Math-1.5B, while OPD uses JustRL-DeepSeek-1.5B as the
teacher and DeepSeek-R1-Distill-Qwen-1.5B as the student.}
\label{tab:obj-agnostic}
\vspace{-2mm}
\end{table}

\subsection{Objective-Agnostic Gains~(RQ2)}
In this section, we show that \Algnameabbr{}'s improvement generalizes to various RL post-training objectives. Recent advancements in RL post-training have proposed different objective designs to improve the stability, efficiency, and effectiveness of policy optimization. 
To evaluate whether \Algnameabbr{} can complement these objective-level advances, we consider three representative recent methods: GSPO, DAPO, and OPD.
As shown in the Table~\ref{tab:obj-agnostic}, \Algnameabbr{} consistently improves all three methods across reasoning benchmarks. These results indicate that the benefit of \Algnameabbr{} is not tied to a specific RL objective. Instead, \Algnameabbr{} improves the general data schedule of RL post-training, where the inter-cluster distribution adapts sampling toward semantic clusters aligned with the policy's current capability, while intra-cluster scheduling selects policy-boundary samples. As a result, different methods can optimize on more informative samples while enhancing the gains from their original objective designs.
\begin{table}[t]
\centering
\small
\setlength{\tabcolsep}{3pt}
\renewcommand{\arraystretch}{1.08}
\begin{tabular}{@{}lrrrr@{}}
\toprule
& \multicolumn{2}{c}{\textbf{Qwen2.5-Math-1.5B}}
& \multicolumn{2}{c}{\textbf{Olmo3-7B-SFT}} \\
\cmidrule(lr){2-3}
\cmidrule(lr){4-5}
Dataset & Own & Transferred & Own & Transferred \\
\midrule
MATH-500 & 63.47 & 63.72 & 67.56 & 65.84 \\
AIME 25  & 7.08  & 7.02  & 18.96 & 18.19 \\
AIME 24  & 9.44  & 10.56 & 18.33 & 16.32 \\
Minerva  & 17.20 & 16.83 & 18.58 & 18.89 \\
AMC 23   & 45.89 & 46.56 & 55.57 & 51.62 \\
GPQA-D   & 27.28 & 26.14 & 17.84 & 18.50 \\
MMLU-Pro & 25.96 & 27.09 & 13.56 & 14.02 \\
\midrule
Average  & 28.05 & 28.27 & 30.06 & 29.05 \\
\bottomrule
\end{tabular}
\caption{Cross-policy clustering robustness of \Algnameabbr{}, comparing policy-specific clustering with clusters
transferred from another policy. We report Mean@16 accuracy (\%).
``Own'' uses the target policy for clustering, while
``Transferred'' uses the other policy.}
\label{tab:cluster-transfer}
\vspace{-2mm}
\end{table}
\subsection{Cross-Policy Clustering Robustness~(RQ3)}
\label{sec:cross-policy}
In this section, we show that \Algnameabbr{} is robust to the policy used for constructing semantic clusters. In the default setting, \Algnameabbr{} constructs clusters using the base policy being trained, which directly aligns the clustering representation with the target policy. To examine whether the clustering step is overly dependent on this specific policy, we conduct a cross-policy clustering study between Qwen2.5-Math-1.5B and OLMo3-7B-Instruct-SFT. Specifically, we use clusters constructed by OLMo3-7B-Instruct-SFT to train Qwen2.5-Math-1.5B, and conversely use clusters constructed by Qwen2.5-Math-1.5B to train OLMo3-7B-Instruct-SFT. We compare these cross-policy variants with the default \Algnameabbr{} setting, where each policy uses its own clusters. As shown in Table~\ref{tab:cluster-transfer}, the cross-policy clustering variants achieve performance comparable to the default setting. This suggests that \Algnameabbr{} is not overly sensitive to the exact policy used for semantic clustering, and that the constructed clusters capture semantic structure shared across different policies. 
These results indicate that cross-policy clustering is feasible, yet the default setting provides the best accuracy-efficiency trade-off: using a larger policy for clustering induces additional computational overhead, while using a smaller policy can cause a mild performance drop.
Therefore, constructing clusters with the policy being trained remains the most practical choice.

\begin{figure*}[t]
    \centering
    \begin{subfigure}[t]{0.48\textwidth}
        \centering
        \includegraphics[width=\linewidth]{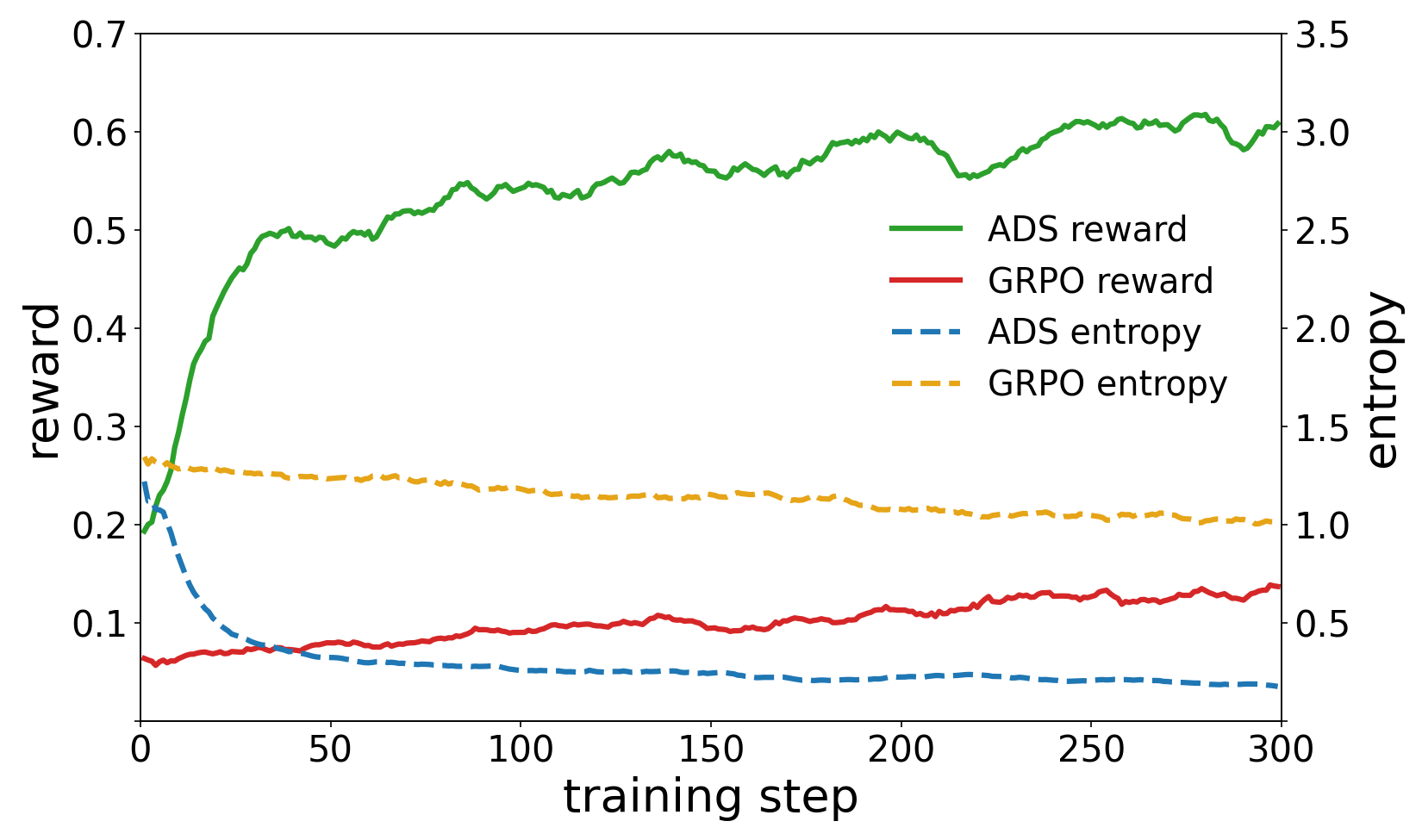}
        \caption{Qwen2.5-Math-1.5B}
        \label{fig:qwen25_math_15b}
    \end{subfigure}
    \hfill
    \begin{subfigure}[t]{0.48\textwidth}
        \centering
        \includegraphics[width=\linewidth]{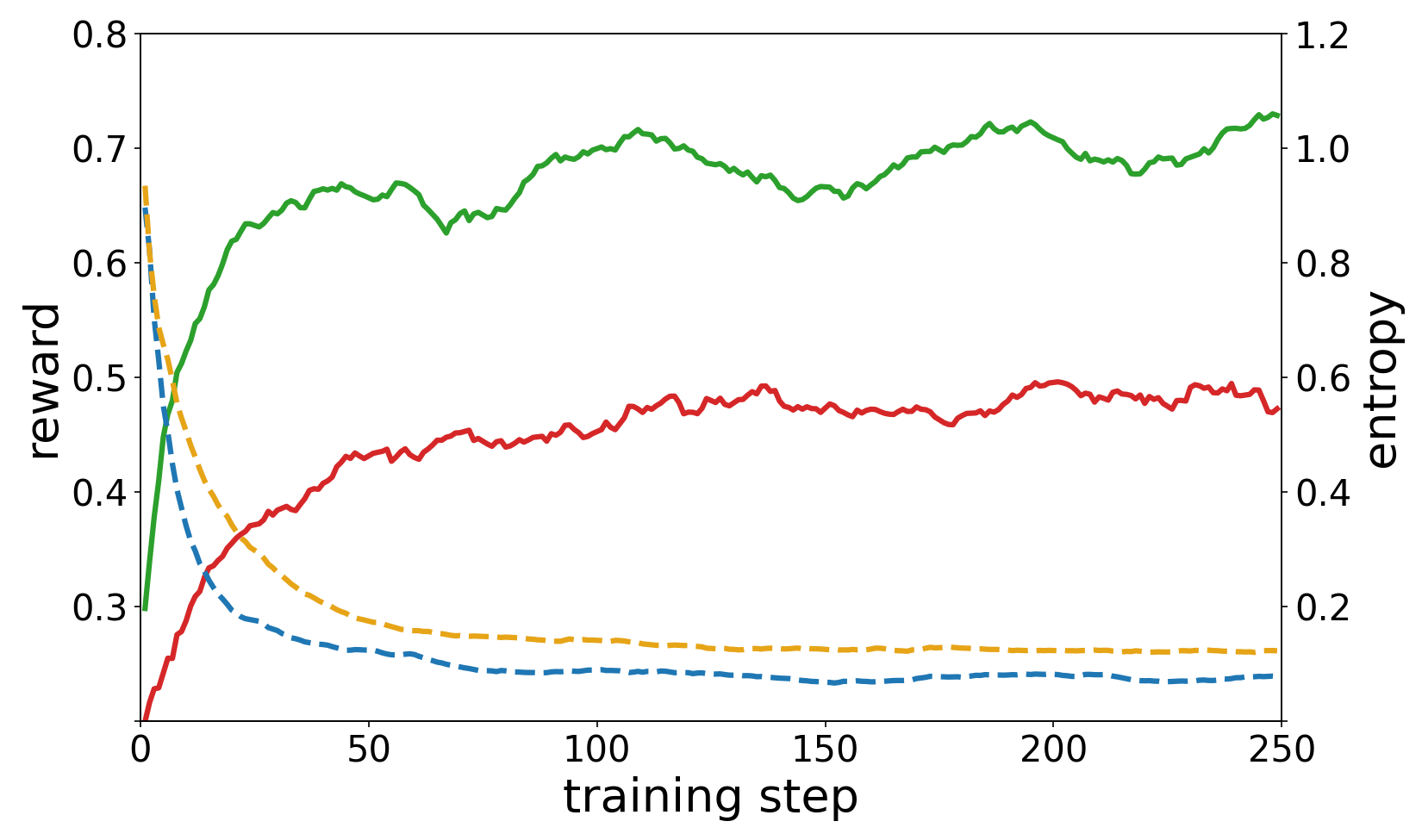}
        \caption{Qwen3-4B-Base}
        \label{fig:qwen3_4b_base}
    \end{subfigure}
    \caption{Training dynamics comparison of \Algnameabbr{} and GRPO. \Algnameabbr{} exhibits a spiral reward increase.}
    \label{fig:reward_entropy_comparison}
    \vspace{-1mm}
\end{figure*}

\begin{table*}[t]
\centering
\footnotesize
\setlength{\tabcolsep}{3pt}
\renewcommand{\arraystretch}{1.08}
\begin{tabular}{@{}lrrrrrrrr@{}}
\toprule
Method & MATH-500 & AIME 25 & AIME 24 & Minerva
& AMC 23 & GPQA-D & MMLU-Pro & Average \\
\midrule
\Algnameabbr{}
& \textbf{63.47} & \textbf{7.08} & 9.44
& \textbf{17.20} & 45.89 & \textbf{27.28}
& \textbf{25.96} & \textbf{28.05} \\
\midrule
without Inter-Cluster Distribution
& 61.78 & 5.07 & 9.65 & 16.28
& 44.84 & 26.16 & 25.75 & 27.08 \\
without Intra-Cluster Scheduling
& 62.19 & 5.62 & 9.65 & 16.12
& 43.80 & 26.31 & 24.89 & 26.94 \\
Global Easy-to-Hard Scheduling
& 61.58 & 4.38 & 9.58 & 15.90
& 44.38 & 26.16 & 25.42 & 26.77 \\
Random Clustering
& 61.92 & 6.88 & 9.38 & 14.43
& \textbf{47.50} & 23.01 & 22.80 & 26.56 \\
Boundary-Targeted Cluster Update
& 62.80 & 6.67 & \textbf{10.00} & 16.18
& 35.00 & 24.24 & 23.80 & 25.53 \\
Uniform Sampling
& 55.04 & 3.47 & 5.63 & 13.63
& 35.68 & 20.14 & 17.37 & 21.57 \\
\bottomrule
\end{tabular}
\caption{Ablations of \Algnameabbr{} on Qwen2.5-Math-1.5B.
We report Mean@16 accuracy (\%).}
\label{tab:ablation}
\vspace{-3mm}
\end{table*}
\subsection{Training Dynamics Analysis}

\Algnameabbr{} produces a paced training reward trajectory that reflects its continuous adaptation to the policy's capability boundary. Figure~\ref{fig:reward_entropy_comparison} shows that the reward curve of \Algnameabbr{} follows a spiral-like pattern with repeated local adjustments, while GRPO exhibits a slower reward increase under uniform sampling. 
This behavior matches our adaptive data scheduling mechanism, where \Algnameabbr{} advances toward harder samples when selected samples become too easy for the current policy, and moves back toward easier neighboring samples when selected samples become too difficult. Therefore, the reward can fluctuate locally while still improving over the long run. Notably, although the reward exhibits local fluctuations, \Algnameabbr{} maintains an improving reward trend and a smoother entropy trajectory than GRPO, suggesting stable policy optimization while targeting policy-boundary samples~\citep{yu2025dapo,liu2025understanding,park2025clip}.

\subsection{Ablation Studies}
In this section, we conduct ablation studies to test the
contributions of each module of \Algnameabbr{}.
Removing the inter-cluster distribution treats the full dataset
as a single cluster for global policy-boundary selection, while
removing intra-cluster scheduling retains adaptive cluster
selection but samples uniformly within each selected cluster.
To assess semantic grouping, random clustering replaces
semantic clusters with random partitions.
We also compare with global easy-to-hard schedule, which
progresses from the easiest prompts to the hardest according
to fixed offline difficulty scores. Furthermore, to validate the inter-cluster update rule, we further replace Eq.~\ref{eq:inter_cluster_probability} with $p_k^{m+1}=(1-2|r_k^m-0.5|)/\sum_{j=1}^{K} r_j^m$, prioritizing
clusters with success rates near $50\%$ while keeping other components fixed.
Table~\ref{tab:ablation} shows that the full \Algnameabbr{}
achieves best average accuracy, indicating that all modules are necessary for the best gains and validating \Algnameabbr{}'s dual-level design.

\section{Related Work}
LLM post-training has advanced through group-level and sequence-level policy optimization~\cite{guo2025deepseek, zheng2025group} and on-policy distillation~\cite{agarwal2024policy}. Data scheduling provides a complementary approach. DAPO~\cite{yu2025dapo} filters rollout groups with uniform rewards, AdaRFT~\cite{adarft} adapts a global difficulty target using reward feedback, and DOTS+RR~\cite{sun2025improving}
combines online difficulty prediction with rollout replay.
However, data scheduling that jointly accounts for the semantic
structure of training data and the policy's evolving capability
remains underexplored. Additional related work is discussed in
Appendix~\ref{app:related_work}.

\section{Conclusion}
In this work, we introduced Adaptive Data Scheduling (\Algnameabbr{}), a dual-level data scheduling framework that enhances RL post-training LLMs. \Algnameabbr{} enables the training policy to learn at an optimal pace by replacing the standard uniform sampling with an adaptive sampling distribution and policy-boundary sample selection. At the cluster level, the framework prioritizes semantic clusters that align with the policy’s current capability to solidify learning progress. At the sample level, it tracks policy-boundary samples within these clusters to ensure each training step provides informative relative advantages. Experimental results across three LLMs and seven benchmarks show that \Algnameabbr{} consistently improves accuracy and generalizes across RL objectives. These results show the effectiveness of \Algnameabbr{} for enhancing RL post-training performance.
\section*{Limitations}
In this work, we proposed \Algnameabbr{} to adaptively select training samples based on semantic structure and policy capability in RL with verifiable outcome reward settings.  \Algnameabbr{} identifies policy-boundary samples through empirical rollout success rates, which naturally fits reasoning tasks with verifiable outcome rewards. Extending this idea to broader post-training settings can be an important future direction. For instance, open-ended instruction following, preference optimization, tool use, and multi-turn interaction may require richer notions of sample informativeness beyond binary correctness. Developing adaptive scheduling signals for these reward settings could make \Algnameabbr{} applicable to a wider range of LLM post-training pipelines. More broadly, our results suggest that data scheduling is a complementary dimension to objective design in RL post-training. Future work can combine \Algnameabbr{} with other RL objectives to further improve the performance and stability of post-training.



\bibliography{references}

\appendix

\section{Boundary Criterion Justification}
\label{sec:5050just}
For a sample prompt $x_i$ at a given training step, let $\bar\rho_i=\Pr_{o\sim\pi_\theta(\cdot\mid x_i)}[R(x_i,o)=1]$ be its success probability under the current policy. The rollout rewards $R_{i,j}:=R(x_i,o_{i,j})$ are then i.i.d.\ $\mathrm{Bernoulli}(\bar\rho_i)$, and their mean $\mu_i$ is the empirical success rate $\rho_i$ of Section~3.3.
 
\begin{proposition}
\label{prop:boundary}
Under binary rewards, (i) the advantages $\{A_{i,j}\}_{j=1}^G$ are not all zero if and only if the group contains both correct and incorrect responses, which occurs with probability
\begin{equation}
\eta_G(\bar\rho_i)=1-\bar\rho_i^{\,G}-(1-\bar\rho_i)^G;
\label{eq:eta}
\end{equation}
and (ii) the expected squared centered advantage is
\begin{equation}
\mathbb{E}\Big[\frac{1}{G}\sum_{j=1}^G (R_{i,j}-\mu_i)^2\Big]=\Big(1-\frac{1}{G}\Big)\bar\rho_i(1-\bar\rho_i).
\label{eq:sq_adv}
\end{equation}
Both are maximized at $\bar\rho_i=1/2$ and vanish as $\bar\rho_i\to 0$ or $1$.
\end{proposition}
 
\begin{proof}
(i) If all rewards are equal, $R_{i,j}=\mu_i$ for every $j$ and all advantages vanish; otherwise $\mu_i\in(0,1)$, so $\sigma_i>0$ and $R_{i,j}-\mu_i\neq0$ for every $j$. By independence, all-correct and all-incorrect groups occur with probabilities $\bar\rho_i^{\,G}$ and $(1-\bar\rho_i)^G$, and $\eta_G$ is strictly concave and symmetric about $1/2$, hence maximized there. (ii) As $R_{i,j}^2=R_{i,j}$, the left-hand side equals $\mathbb{E}[\mu_i-\mu_i^2]$. Substituting $\mathbb{E}[\mu_i]=\bar\rho_i$ and $\mathbb{E}[\mu_i^2]=\bar\rho_i^{\,2}+\bar\rho_i(1-\bar\rho_i)/G$ gives Eq.~\eqref{eq:sq_adv}.
\end{proof}
 
Since Eq.~\eqref{eq:sq_adv} is proportional to the expected squared norm of the GRPO gradient estimate \citep[Theorem~1]{sun2025improving}, samples with $\bar\rho_i$ near $1/2$ yield the most informative updates.

\section{Offline Preprocessing Cost}

\Algnameabbr{} includes an offline preprocessing stage before RL training, where we compute sample embeddings for semantic clustering and estimate offline difficulty scores for intra-cluster scheduling. This preprocessing stage is performed only once before training starts. After the semantic clusters and difficulty ordering are constructed, they are reused throughout the entire RL post-training process and do not introduce additional per-step overhead during policy optimization. Table~\ref{tab:offline_preprocessing_cost} reports the wall-clock cost of this offline stage. To contextualize the cost relative to RL training, we also report the training-step equivalent, computed as the offline preprocessing time divided by the average wall-clock time of one RL training step for the corresponding model. Across all evaluated models, the offline stage corresponds to only a small number of RL training steps. This indicates that the preprocessing cost is minimal compared with the total cost of RL post-training, especially since online training repeatedly generates multiple long rollouts per prompt while the offline stage is incurred only once. Moreover, the offline cost can be further reduced in practice. As shown in Section~\ref{sec:cross-policy}, \Algnameabbr{} is robust to cross-policy clustering, where clusters constructed from one policy can be used for another policy with comparable performance. This suggests that the semantic clustering stage does not always need to be recomputed for every target policy. In settings where multiple related policies are trained on the same data, reusable clusters can further reduce the offline preprocessing cost.

\begin{table}[h]
\centering
\small
\setlength{\tabcolsep}{8pt}
\renewcommand{\arraystretch}{1.08}
\begin{tabular}{lcc}
\toprule
Model &  Time & Step Equivalent \\
\midrule
Qwen2.5-Math-1.5B & 15.5 min & $\approx$ 6 steps \\
Qwen3-4B-Base     & 21.5 min & $\approx$ 9 steps \\
OLMo3-7B-SFT      & 26.6 min & $\approx$ 3 steps \\
\bottomrule
\end{tabular}
\caption{Offline preprocessing cost of \Algnameabbr{}. The training-step equivalent is computed by dividing the offline preprocessing time by the average wall-clock time of one RL training step for the policy.}
\label{tab:offline_preprocessing_cost}
\end{table}

\section{The Algorithm of \Algnameabbr{}}
\label{sec:algo}
Algorithm~\ref{alg:ads} summarizes the training procedure of \Algnameabbr{}.
The algorithm begins by organizing the dataset into semantic clusters and sorting each cluster by sample difficulty before training (line~1-3). The boundary mini-cluster is initialized with the easiest samples in each cluster and the inter-cluster distribution is initialized uniformly (line~4-7). 
During training, \Algnameabbr{} samples clusters according to the current inter-cluster distribution and updates the policy using all samples in their boundary mini-cluster (line~8-10).
Based on the rollout outcomes, \Algnameabbr{} calculates both sample-level and cluster-level success rate to update the boundary mini-cluster and inter-cluster distribution (line~11-14). Once all samples in a cluster have been scheduled at least once, the cluster is removed from $\mathcal{C}$ until the next epoch to encourage coverage across clusters, according to the cluster traversal rule (line~14-16).

\begin{algorithm}[t]
\caption{Adaptive Data Scheduling (\Algnameabbr{})}
\label{alg:ads}
\begin{algorithmic}[1]
\REQUIRE Dataset $\mathcal{D}$, base policy $\pi_\theta$, number of clusters $K$, mini-cluster size $B$
\STATE Compute $\{\mathbf{e}_i\}_{i=1}^{N}$ by Eq.~\ref{eq:embedding}
\STATE Construct $\mathcal{C}=\{C_1,\ldots,C_K\}$ by Eq.~\ref{eq:semantic_clustering}
\STATE Compute difficulty scores $\{d_i\}_{i=1}^{N}$ by Eq.~\ref{eq:difficulty}
\FOR{$k=1,\ldots,K$}
    \STATE $C_k \leftarrow \mathrm{sort}(C_k; d_i \uparrow)$
    \STATE $M_k^0 \leftarrow C_k[1:B]$, \quad $p_k^0 \leftarrow 1/K$
\ENDFOR
\FOR{training step $m=0,1,2,\ldots$}
    \STATE Sample clusters $C_k \sim \{p_k^{m}\}_{k=1}^{K}$
    \STATE Update $\pi_\theta$ with $(x_a, y_a)\in M_k^m$
    \STATE Compute $\rho_a^m$ and $r_k^m$
    \STATE Update $M_k^{m+1}$ by Eq.~\ref{eq:mini_cluster_update}
    \STATE Update $\{p_k^{m+1}\}_{k=1}^{K}$ by Eq.~\ref{eq:inter_cluster_probability}
    \IF{all samples in $C_k$ are scheduled once}
        \STATE $\mathcal{C} \leftarrow \mathcal{C} \setminus \{C_k\}$ until the next epoch
    \ENDIF
\ENDFOR
\end{algorithmic}
\end{algorithm}

\begin{table}[t]
\centering
\small
\begin{tabular}{lcccc}
\toprule
 & & \multicolumn{3}{c}{ADS ($\epsilon$)} \\
\cmidrule(lr){3-5}
Dataset & GRPO & 0.05 & 0.17$^*$ & 0.25 \\
\midrule
MATH-500 & 55.04 & 63.29 & \textbf{63.47} & 62.48 \\
AIME 25 & 3.47 & 6.39 & \textbf{7.08} & 6.88 \\
AIME 24 & 5.63 & \textbf{9.51} & 9.44 & 8.12 \\
Minerva & 13.63 & 16.44 & \textbf{17.20} & 16.08 \\
AMC 23 & 35.68 & 45.10 & 45.89 & \textbf{47.66} \\
GPQA-D & 20.14 & 25.69 & \textbf{27.28} & 25.32 \\
MMLU-Pro & 17.37 & \textbf{26.01} & 25.96 & 25.07 \\
\midrule
Average & 21.57 & 27.49 & \textbf{28.05} & 27.37 \\
\bottomrule
\end{tabular}
\caption{Sensitivity analysis of ADS to the tolerance band width $\epsilon$.  We report Mean@16 accuracy (\%).}
\label{tab:sens_eps}
\end{table}
 
\begin{table}[t]
\centering
\small
\begin{tabular}{lcccc}
\toprule
 & & \multicolumn{3}{c}{ADS ($\alpha$)} \\
\cmidrule(lr){3-5}
Dataset & GRPO & 0.2 & 0.3$^*$ & 0.4 \\
\midrule
MATH-500 & 55.04 & 62.72 & \textbf{63.47} & 62.84 \\
AIME 25 & 3.47 & 6.33 & 7.08 & \textbf{8.12} \\
AIME 24 & 5.63 & \textbf{9.48} & 9.44 & 8.75 \\
Minerva & 13.63 & 16.35 & \textbf{17.20} & 16.75 \\
AMC 23 & 35.68 & 44.84 & \textbf{45.89} & 45.47 \\
GPQA-D & 20.14 & 24.43 & \textbf{27.28} & 26.64 \\
MMLU-Pro & 17.37 & 25.57 & \textbf{25.96} & 25.20 \\
\midrule
Average & 21.57 & 27.10 & \textbf{28.05} & 27.68 \\
\bottomrule
\end{tabular}
\caption{Sensitivity analysis of ADS to the smoothing coefficient $\alpha$. We report Mean@16 accuracy (\%).}
\label{tab:sens_alpha}
\end{table}

\section{More on Implementation Details}
\label{sec:implementation_details}
As discussed in Section~\ref{sec:inter-cluster}, we employ exponential smoothing when updating the inter-cluster distribution. Specifically, we use the calculated $\hat{p}_k^{m+1} = r_k^m/\sum_{j=1}^{K} r_j^m$ as a target probability and applies exponential smoothing $p_k^{m+1} = (1-\alpha)p_k^m+\alpha \hat{p}_k^{m+1}$, where $\alpha$ controls the adaptation rate. In all experiments, we set  $\alpha=0.3$ to avoid sudden shifts in the inter-cluster distribution caused by noisy rollouts. For intra-cluster scheduling, we set the mini-cluster size to $B=32$. Since the rollout batch size is $128$, each training step samples $4$ semantic clusters according to the current inter-cluster distribution and uses the $32$ samples in the boundary mini-cluster of each selected cluster. This gives a total of $128$ policy-boundary samples for rollout generation and policy optimization at each step. We also reinitialize the scheduler state at the beginning of each new training epoch. Specifically, we use the same epoch length as standard GRPO-style training: if the training set contains $|\mathcal{D}|$ prompts and the batch size is $b$, a new epoch starts after $ |\mathcal{D}|/b$ update steps. At this epoch boundary, we reset the inter-cluster distribution and reinitialize each boundary mini-cluster. This reset mechanism is to prevent rollout statistics from earlier policy states from dominating later scheduling decisions, while still allowing \Algnameabbr{} to adapt within each epoch to the current capability of the policy.

\section{Evaluation Details}
\label{app:eval_detail}
For all evaluation, we use the vLLM library~\cite{vllm}, with sampling temperature set to $0.7$, top-$p$ to 0.95, and max generation length set to 16{,}382, with the answer extraction prompt of ``Please reason step by step, and put your final answer within \texttt{\textbackslash boxed\{\}}'', following exisiting works~\cite{xu2025dts, chen2025verithinker, luo2026demystifying}. For mathematical benchmarks, we compare the final boxed answer with the reference by normalized exact match or numerical equality. For GPQA-D and MMLU-Pro, a response is counted as correct only if it commits to the correct option letter, e.g., \texttt{\textbackslash boxed\{B\}}. This format-sensitive protocol is applied identically to all methods to assess both instruction following and reasoning performance. 

\begin{figure*}[t]
    \small
    \centering
    \includegraphics[width=1\linewidth]{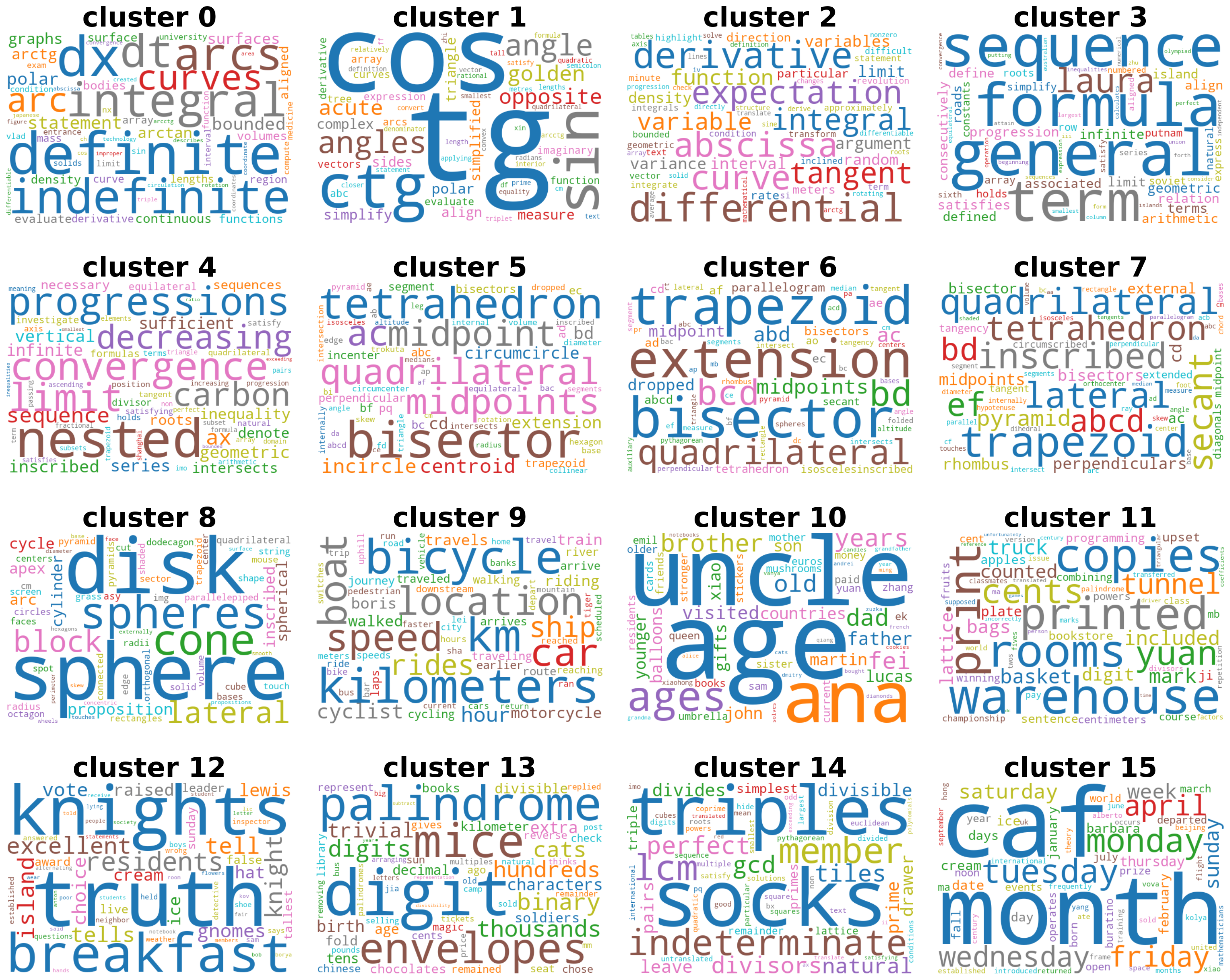}
    \caption{Word cloud visualization of 16 representative semantic clusters constructed from Qwen2.5-Math-1.5B embeddings, showing coherent domains such as calculus, geometry, sequences, counting, age reasoning, and calendar reasoning.}
    \label{fig:wordcloud}
\end{figure*}

\begin{table}[t]
\centering
\small
\begin{tabular}{lcccc}
\toprule
 & & \multicolumn{3}{c}{ADS ($K$)} \\
\cmidrule(lr){3-5}
Dataset & GRPO & 32 & 64$^*$ & 128 \\
\midrule
MATH-500 & 55.04 & 62.21 & \textbf{63.47} & 62.58 \\
AIME 25 & 3.47 & 6.67 & \textbf{7.08} & 4.58 \\
AIME 24 & 5.63 & 10.83 & 9.44 & \textbf{11.25} \\
Minerva & 13.63 & 16.38 & \textbf{17.20} & 16.31 \\
AMC 23 & 35.68 & 42.50 & 45.89 & \textbf{47.97} \\
GPQA-D & 20.14 & 25.85 & \textbf{27.28} & 25.13 \\
MMLU-Pro & 17.37 & 25.44 & \textbf{25.96} & 24.50 \\
\midrule
Average & 21.57 & 27.13 & \textbf{28.05} & 27.47 \\
\bottomrule
\end{tabular}
\caption{Sensitivity analysis of ADS to the number of clusters $K$. We report Mean@16 accuracy (\%).}
\label{tab:sens_K}
\end{table}
 
\begin{table}[t]
\centering
\small
\begin{tabular}{lcccc}
\toprule
 & & \multicolumn{3}{c}{ADS ($B$)} \\
\cmidrule(lr){3-5}
Dataset & GRPO & 16 & 32$^*$ & 64 \\
\midrule
MATH-500 & 55.04 & 62.88 & 63.47 & \textbf{63.52} \\
AIME 25 & 3.47 & \textbf{7.71} & 7.08 & 6.25 \\
AIME 24 & 5.63 & \textbf{10.35} & 9.44 & 9.38 \\
Minerva & 13.63 & 16.51 & \textbf{17.20} & 16.59 \\
AMC 23 & 35.68 & 45.47 & 45.89 & \textbf{47.03} \\
GPQA-D & 20.14 & 25.63 & \textbf{27.28} & 26.14 \\
MMLU-Pro & 17.37 & 25.59 & \textbf{25.96} & 25.86 \\
\midrule
Average & 21.57 & 27.73 & \textbf{28.05} & 27.82 \\
\bottomrule
\end{tabular}
\caption{Sensitivity analysis of ADS to the mini-cluster size $B$. We report Mean@16 accuracy (\%).}
\label{tab:sens_B}
\end{table}

\section{Controlled Scaling within Qwen3}
\label{sec:scaling}
To assess performance across model sizes while controlling model family and initial training stage, we compare GRPO and \Algnameabbr{} on Qwen3-Base models at 1.7B, 4B, and 8B~\cite{yang2025qwen3}. We report Mean@16 accuracy on the same seven benchmarks used in the main experiments. As shown in Table~\ref{tab:qwen3_scaling}, the gains persist across the tested scales within a single model family.

\section{Sensitivity Analysis}
We evaluate the sensitivity of ADS to its four hyperparameters on Qwen2.5-Math-1.5B: the tolerance band width $\epsilon$, the smoothing coefficient $\alpha$, the number of clusters $K$, and the mini-cluster size $B$. We vary one hyperparameter at a time while keeping the others at their defaults. As shown in Tables~\ref{tab:sens_eps}--\ref{tab:sens_B}, across all tested values, \Algnameabbr{} remains stable and consistently outperforms GRPO, indicating that the method is robust and relatively insensitive to hyperparameter setting.

\begin{table}[t]
    \centering
    \small
    \setlength{\tabcolsep}{3pt}
    \caption{Controlled scaling comparison in the Qwen3-Base family. We report Mean@16 accuracy (\%).}
    \label{tab:qwen3_scaling}
    \begin{tabular}{lrrrrrr}
        \toprule
        & \multicolumn{2}{c}{1.7B}
        & \multicolumn{2}{c}{4B}
        & \multicolumn{2}{c}{8B} \\
        \cmidrule(lr){2-3}\cmidrule(lr){4-5}\cmidrule(lr){6-7}
        Dataset & GRPO & \Algnameabbr{} & GRPO & \Algnameabbr{} & GRPO & \Algnameabbr{} \\
        \midrule
        MATH-500 & 51.29 & \textbf{58.13} & 61.49 & \textbf{67.13} & 67.53 & \textbf{70.60} \\
        AIME25 & 2.92 & \textbf{6.46} & 11.32 & \textbf{11.53} & 12.08 & \textbf{15.62} \\
        AIME24 & 4.58 & \textbf{7.08} & 12.22 & \textbf{12.50} & 13.96 & \textbf{18.12} \\
        Minerva & 9.17 & \textbf{14.52} & 18.40 & \textbf{20.89} & 21.12 & \textbf{23.39} \\
        AMC23 & 30.63 & \textbf{35.31} & 53.65 & \textbf{54.89} & 56.56 & \textbf{64.06} \\
        GPQA-D & 11.08 & \textbf{11.68} & 16.11 & \textbf{20.69} & 16.95 & \textbf{17.61} \\
        MMLU-Pro & \textbf{16.14} & 16.05 & 20.04 & \textbf{20.96} & 20.47 & \textbf{20.67} \\
        \midrule
        Average & 17.97 & \textbf{21.32} & 27.60 & \textbf{29.80} & 29.81 & \textbf{32.87} \\
        \bottomrule
    \end{tabular}
\end{table}

\section{Semantic Clusters Visualization}
\label{app:wordcloud}
To qualitatively examine whether the semantic clustering module captures meaningful structure in the training data, we visualize 16 representative clusters out of 64 clusters generated from Qwen2.5-Math-1.5B using word clouds in Figure~\ref{fig:wordcloud}. For each cluster, we display the most frequent tokens after basic text normalization. The visualization shows that the learned clusters correspond to coherent semantic patterns rather than arbitrary partitions of the dataset. Several clusters align with recognizable mathematical domains. For example, some clusters focus on calculus-related concepts such as derivatives, integrals, limits, convergence, and indefinite forms. There also exist geometric-focused domains, containing terms related to quadrilaterals, trapezoids, bisectors, tetrahedrons, spheres, and circles. These patterns suggest that the representation space clustering samples according to shared semantic and structural properties and successfully identified semantically coherent clusters, which supports the use of cluster-level and sample-level scheduling in \Algnameabbr{}.

\section{Additional Related Work}
\label{app:related_work}
\paragraph{RL for Post-Training.}
Reinforcement learning has become an important paradigm for post-training large language models. 
Early Reinforcement Learning from Human Feedback (RLHF) pipelines commonly adopt PPO~\cite{schulman2017proximal} to optimize language models with reward feedback from sampled responses. 
To simplify preference-based post-training, \cite{rafailov2023direct} proposes Direct Preference Optimization (DPO), which directly optimizes policies from preference data without explicitly training a reward model or performing online policy optimization. 
For reasoning tasks with verifiable rewards, GRPO~\cite{guo2025deepseek} has emerged as a widely used framework by replacing the value model in PPO with group-relative advantage estimation over multiple rollouts from the same prompt.
Recent methods further extend GRPO-style post-training through improved objectives, supervision signals, and optimization strategies. 
For instance, DAPO~\cite{yu2025dapo} refines long-chain reasoning optimization, GSPO~\cite{zheng2025group} performs sequence-level policy optimization, and OPD~\cite{agarwal2024policy} introduces teacher-guided on-policy distillation. 
While these advances improve policy optimization, they leave the training prompt schedule largely underexplored.
Despite their different designs, these methods typically rely on uniform prompt sampling, treating the training set as an unstructured pool. 
Such a static schedule ignores the semantic structure of training data and the evolving capability of the policy, which can limit the effectiveness of RL updates. 
Together, these limitations suggest the need for a structured data schedule that adapts to both the training data and the current policy state. 
\Algnameabbr{} addresses this gap by keeping the underlying RL objective unchanged and adaptively scheduling prompts according to semantic structure and policy capability.

\begin{table}[t]
\centering
\begin{tabular}{l|c}
\toprule
Name & Value \\
\midrule
Data type & \texttt{torch.bfloat16} \\
Flash-Attention & True \\
Computing Infrastructure & GPU \\
GPU Model & NVIDIA-H200 \\ 
GPU Memory & 141 GB \\ 
GPU Number & 4 \\
CUDA Version & 12.9 \\
CPU Memory & 512GB \\
\bottomrule
\end{tabular}
\caption{Experiment configuration and computing infrastructure.}
\label{tab:computing_infrastructure}
\end{table}

\paragraph{Data Scheduling.}
Data scheduling has been widely studied as a way to improve optimization by changing the effective training distribution over time. 
In general machine learning and reinforcement learning, samples are not equally useful throughout training: some may be redundant for the current model, while others may provide stronger learning signals. 
Adaptive scheduling therefore adjusts which samples are used at different stages, improving learning efficacy without changing the model architecture or optimization objective.
Recent works~\cite{yu2025dapo,adarft,fastcurl,sun2025improving} have aimed to apply this idea to LLM reinforcement learning. 
However, existing approaches mainly perform prompt-level selection and still view the dataset as a flat pool, overlooking semantic structure and training progress across different semantic regions. 
To address these limitations, \Algnameabbr{} schedules data at two levels: it adapts sampling across semantic clusters and then selects policy-boundary samples within each cluster. 

\section{Computational Infrastructure}
\label{sec:app-computing}
The computational infrastructure information is given in Table~\ref{tab:computing_infrastructure}.

\section{Potential Risks}
All open-source packages used in this work have licenses that permit academic research. We use and cite public benchmarks and open models strictly for research and evaluation, in accordance with their original licenses and terms of use. To our knowledge, these datasets contain no personally identifying information or offensive content. ADS only changes how training data is sampled and does not introduce capabilities beyond those of the underlying RL objective, so we do not foresee risks beyond those of the base models and post-training methods it builds on. AI assistants were used to help refine the writing and to assist with implementing the proposed algorithm, and the authors reviewed all content.

\end{document}